\pdfoutput=1

\documentclass[11pt]{article}

\usepackage[]{acl}

\usepackage{times}
\usepackage{latexsym}

\usepackage[T1]{fontenc}

\usepackage[utf8]{inputenc}

\usepackage{microtype}

\usepackage{mwe}
\usepackage{amssymb}
\usepackage{amsfonts}
\usepackage{amsmath}
\usepackage{mathtools}
\usepackage{tikz}
\usepackage{pgfplots}
\usepackage{tikz-qtree}
\usepackage{tikz-dependency}
\usepackage{forest}
\usepackage{multirow}
\usepackage{subcaption}
\usepackage{booktabs}
\usepackage{proof}
\usepackage{hyperref}
\usepackage{csquotes}
\usepackage[linesnumbered,ruled,lined]{algorithm2e}
\usepackage{bm}
\usepackage{arydshln}
\useforestlibrary{linguistics}

\title{Nested Named Entity Recognition as Latent Lexicalized Constituency Parsing}

\author{Chao Lou, Songlin Yang, Kewei Tu\thanks{\; Corresponding Author}\\
  School of Information Science and Technology, ShanghaiTech University \\
  Shanghai Engineering Research Center of Intelligent Vision and Imaging\\ 
    {\tt \{louchao,yangsl,tukw\}@shanghaitech.edu.cn}\\
 }

\pgfplotsset{compat=1.17}
\begin{document}
\maketitle
\begin{abstract}
Nested named entity recognition (NER) has been receiving increasing attention. 
Recently, \citet{fu2020nested} adapt a span-based constituency parser to tackle nested NER. They treat nested entities as partially-observed constituency trees and propose the masked inside algorithm for partial marginalization. However, their method cannot leverage entity heads, which have been shown useful in entity mention detection and entity typing. 
In this work, we resort to more expressive structures, lexicalized constituency trees in which constituents are annotated by headwords, to model nested entities. 
We leverage the Eisner-Satta algorithm to perform partial marginalization and inference efficiently.
In addition, we propose to use (1) a two-stage strategy (2) a head regularization loss and (3) a head-aware labeling loss in order to enhance the performance. We make a thorough ablation study to investigate the functionality of each component. Experimentally, our method achieves the state-of-the-art performance on ACE2004, ACE2005 and NNE, and competitive performance on GENIA, and meanwhile has a fast inference speed. Our code will be publicly available at: \href{https://github.com/LouChao98/nner_as_parsing}{github.com/LouChao98/nner\_as\_parsing}.

\end{abstract}

\section{Introduction}\label{sec:introduction}

Named Entity Recognition (NER) is a fundamental task in information extraction, playing an essential role in many downstream tasks. Nested NER brings more flexibility than flat NER by allowing nested structures, thereby enabling more fine-grained meaning representations and broader applications \cite{4338398,dai-2018-recognizing}.  Traditional sequence-labeling-based models have achieved remarkable performance on flat NER but fail to handle nested entities. To resolve this problem, there are many layer-based methods \cite{ju-etal-2018-neural,fisher-vlachos-2019-merge,shibuya-hovy-2020-nested,wang-etal-2020-pyramid, wang-etal-2021-nested} proposed to recognize entities layer-by-layer in bottom-up or top-down manners. However, they suffer from the error propagation issue due to the cascade decoding.

Recently, \citet{fu2020nested} adapt a span-based constituency parser to tackle nested NER, treating annotated entity spans as a partially-observed constituency tree and marginalizing latent spans out for training. Their parsing-based method, namely PO-TreeCRF, admits global exact inference thanks to the CYK algorithm \cite{Cocke1969ProgrammingLA,Younger1967RecognitionAP,Kasami1965AnER}, thereby eliminating the error propagation problem.
However, their method does not consider entity heads,
which provide important clues for entity mention detection \cite{lin-etal-2019-sequence, zhang-etal-2020-two} and entity typing \cite{katiyar-cardie-2018-nested,choi-etal-2018-ultra,chen-etal-2021-explicitly}. For example, \textit{University} and \textit{California} are strong clues of the existence of \texttt{ORGEDU} and \texttt{STATE} entities in Fig.\ref{fig:demo_sent}. Motivated by this and inspired by head-driven phrase structures, \citet{lin-etal-2019-sequence} propose the Anchor-Region Network (ARN), which identifies all entity heads firstly and then predicts the boundary and type of entities governed by each entity head. However, their method is heuristic and greedy, suffering from the error propagation problem as well.

\newcommand{\hw}[1]{{[#1]}}
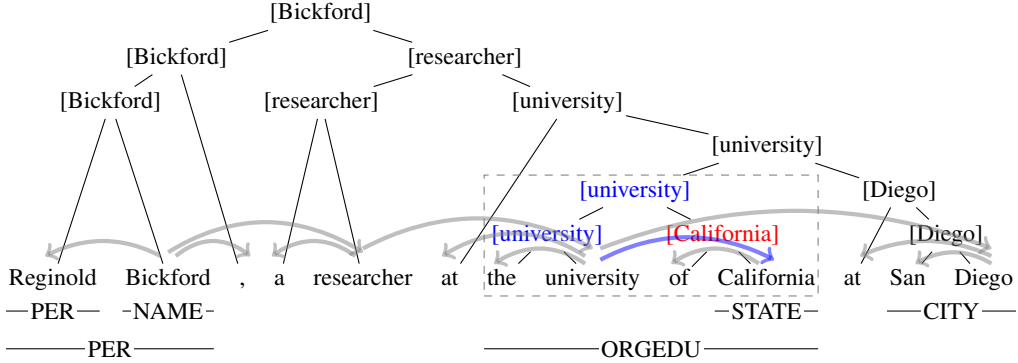
\begin{figure*}[htb!]
  \centering
  \small
  \pgfkeys{/pgf/inner sep=0.1em}
  \pgfplotsset{compat=1.17}

  % ...
  \begin{forest}
    delay={where n children=0{tier=word,
        if={instr("P",content("!u"))}{roof}{}
      }{}},    
    for tree={%
      s sep=0.3cm,
      l sep=0.5em,
      l=1em,
      align=center, base=bottom
    },
    [\hw{Bickford},baseline[\hw{Bickford}[\hw{Bickford}[Reginold,name=r0][Bickford,name=b1
      ]][{,},name=p7]]
    [\hw{researcher}[\hw{researcher}[a,name=a5] [researcher,name=r6]]
        [\hw{university}, before drawing tree={x-=1.7em} [at,name=at1]
            [\hw{university}
              [\hw{university},text=blue,tikz={\node [draw,gray,dashed,inner sep=0,fit to=tree]{};} [\hw{university},text=blue[the,name=t4] [university,name=university]] 
                  [\hw{California},text=red[of,name=of][California,name=c2]]]
              [\hw{Diego}[at,name=at2][\hw{Diego}[San,name=s3] [Diego,name=d4]]]]]]]
    \path (current bounding box.south) coordinate (s1);
    \draw[transform canvas={yshift=-0.2cm}] (r0.south west) -- node[midway,fill=white] {PER} (r0.south east);
    \draw[transform canvas={yshift=-0.2cm}] (b1.south west) -- node[midway,fill=white] {NAME} (b1.south east);
    \draw[transform canvas={yshift=-0.2cm}] (c2.south west) -- node[midway,fill=white] {STATE} (c2.south east);
    \draw[transform canvas={yshift=-0.2cm}] (s3.south west) -- node[midway,fill=white] {CITY} (d4.south east);
    \draw[transform canvas={yshift=-0.7cm}] (r0.south west) -- node[midway,fill=white] {PER} (b1.south east);
    \draw[transform canvas={yshift=-0.7cm}] (t4.south west) -- node[midway,fill=white] {ORGEDU} (c2.south east);
    \draw[->,transform canvas={yshift=+0.1cm,xshift=-0.1cm},gray,ultra thick,opacity=0.5](b1.north) parabola [parabola height=0.2cm] (r0.north);
    \draw[->,transform canvas={yshift=+0.1cm,xshift=+0.1cm},gray,ultra thick,opacity=0.5](b1.north) parabola [parabola height=0.2cm] (p7.north);
    \draw[->,transform canvas={yshift=+0.15cm},gray,ultra thick,opacity=0.5](b1.north) parabola [parabola height=0.4cm] (r6.north);
    \draw[->,transform canvas={yshift=+0.1cm,xshift=-0.1cm},gray,ultra thick,opacity=0.5](r6.north) parabola [parabola height=0.2cm] (a5.north);
    \draw[->,transform canvas={yshift=+0.2cm,xshift=0cm},gray,ultra thick,opacity=0.5](r6.north) parabola [parabola height=0.4cm] (university.north);
    \draw[->,transform canvas={yshift=+0.cm,xshift=-0.1cm},gray,ultra thick,opacity=0.5](university.north) parabola [parabola height=0.2cm] (t4.north);
    \draw[->,transform canvas={yshift=+0.1cm,xshift=-0.1cm},gray,ultra thick,opacity=0.5](university.north) parabola [parabola height=0.3cm] (at1.north);
    \draw[->,transform canvas={yshift=+0.05cm,xshift=+0.1cm},blue,ultra thick,opacity=0.5](university.north) parabola [parabola height=0.3cm] (c2.north);
    \draw[->,transform canvas={yshift=+0.2cm,xshift=+0.1cm},gray,ultra thick,opacity=0.5](university.north) parabola [parabola height=0.5cm] (d4.north);
    \draw[->,transform canvas={yshift=+0.cm,xshift=-0.1cm},gray,ultra thick,opacity=0.5](c2.north) parabola [parabola height=0.2cm] (of.north);
    \draw[->,transform canvas={yshift=+0.cm,xshift=+0.1cm},gray,ultra thick,opacity=0.5](d4.north) parabola [parabola height=0.15cm] (s3.north);
    \draw[->,transform canvas={yshift=+0.1cm,xshift=+0.1cm},gray,ultra thick,opacity=0.5](d4.north) parabola [parabola height=0.2cm] (at2.north);
  \end{forest}
  \setlength{\abovecaptionskip}{24pt}
  \caption{An example sentence with a compatible latent lexicalized constituency tree (top) and observed entities (down). All constituents are annotated by headwords with $[\,\cdot\,]$ and we omit the constituent labels. The dotted frame shows an example of inherited head (blue) and non-inherited head (red). We can draw a dependency arc from the inherited head to the non-inherited head. For example, $\textit{University} \rightarrow \textit{California}$. Hence a lexicalized constituency tree embeds a constituency tree and a dependency tree.}
  \label{fig:demo_sent}
\end{figure*}
Our main goal in this work is to obtain the best of two worlds: proposing a probabilistically principled method that enables exact global inference like \citet{fu2020nested}, meanwhile taking entity heads into accounts like \citet{lin-etal-2019-sequence}. To enable exact global inference, we also view observed entities as partially-observed trees.  Since constituency trees cannot model entity heads, we resort to lexicalized trees, in which constituents are annotated with headwords.  A lexicalized tree embeds a constituency tree and a dependency tree \cite{DBLP:journals/iandc/Gaifman65}, and lexicalized constituency parsing can thus be viewed as joint dependency and constituency parsing \cite{eisner-satta-1999-efficient, collins-2003-head}. Fig.\ref{fig:demo_sent} illustrates an example lexicalized tree. Joint dependency and constituency parsing has been shown to outperform standalone constituency parsing \cite{zhou-zhao-2019-head, DBLP:journals/corr/abs-2009-09730} possibly because modeling dependencies between headwords helps predict constituents correctly.  Hence, in the context of nested NER, we have reasons to believe that modeling latent lexicalized constituency trees would bring improvement in predicting entities over modeling latent constituency trees, and we verify this in experiments.


When using a lexicalized constituency tree for nested NER, only part of unlexicalized spans, i.e., entities, are observed, so we need to marginalize latent spans and dependency arcs out for training.  Inspired by the masked inside algorithm of  \citet{fu2020nested}, we propose a masked version of the Eisner-Satta algorithm \cite{eisner-satta-1999-efficient}, a fast lexicalized constituency parsing algorithm, to perform partial marginalization. We also adopt the Eisner-Satta algorithm for fast inference.

Besides the difference in parsing formalism and algorithms, our work also differs from the work of \citet{fu2020nested} and \citet{lin-etal-2019-sequence} in the following three aspects. First, inspired by \citet{zhang-etal-2020-fast}, we adopt a two-stage parsing strategy, i.e., we first predict an unlabeled tree and then label the predicted constituents, instead of using the one-stage parsing strategy of PO-TreeCRF. We show that two-stage parsing can improve the performance of both PO-TreeCRF and our proposed method. Second, \citet{lin-etal-2019-sequence} observe that each entity head governs only one entity span in most cases, so they impose a hard constraint of that during learning and inference, which is potentially harmful since the constraint is not always satisfied. Instead, we add a soft KL penalty term to encourage satisfaction of the constraint, which is reminiscent of posterior regularization \cite{ganchev_posterior_2010, zhang-etal-2017-prior}. Third,
considering that gold entity heads are not given, \citet{lin-etal-2019-sequence} propose a ``bag loss'' for entity boundary detection and labeling. However, this loss is heuristic and brings an additional hyperparameter, to which the final performance is sensitive. In contrast, entity boundary detection is learned in the first stage of our method, and in the second stage, we propose a more principled labeling loss based on expectations (i.e., marginal likelihoods) of all possible entity heads within gold entity spans, which can be estimated efficiently and does not introduce new hyperparameters.

We conduct experiments on four benchmark datasets, showing that our model achieves state-of-the-art results on ACE2004, ACE2005 and NNE, and competitive results on GENIA, validating the effectiveness of our method.


\section{Preliminary}
\subsection{One-stage and Two-stage Parsing}
A labeled constituency tree can be represented as a rank-3 binary tensor $T$ where $T_{ijk} = 1$ if there is a span from the $i$-th word to the $j$-th word with label $k$ in the tree and $T_{ijk} = 0$ otherwise. We assume the $0$-th label is preserved for $\emptyset$ (i.e., no label) without loss of generality. Similarly, an unlabeled constituency tree can be represented as a binary matrix $T^{\prime}$. 
One-stage span-based constituency parsers decompose the score of a labeled constituency tree into the scores of constituents $s_{i j k}$: \[
 s(T)=\sum_{i j k} T_{i j k} s_{i j k}
\]
They use the CYK algorithm to recover the optimal \textbf{labeled} tree. In contrast, two-stage constituency parsers score unlabeled trees and constituent labels independently. They decompose the score of an unlabeled constituency tree into the scores of spans $s_{i, j}$:
\[
s(T^{\prime}) = \sum_{i j} T^{\prime}_{i j} s_{i j}
\]
They use the CYK algorithm to recover the optimal \textbf{unlabeled} tree in the first stage and then use a separate component to label spans, including the $\emptyset$ label, in the second stage. \citet{Zhang_2020} show that adopting the two-stage parsing strategy leads to a better result in constituency parsing. 

\subsection{PO-TreeCRF}
PO-TreeCRF \cite{fu2020nested} adapts a one-stage constituency parser to tackle nested NER. It views the set of entities  $\bm{y}:=\{(i,j,k), \dots\}$  as observed parts of a constituency tree $T$ where $(i, j)$ is the unlabeled entity span and $k$ is the entity label. We refer to other constituents as latent spans.
A labeled tree $T$ is compatible with $\bm{y}$ if $T_{ijk}=1$ for any entity $(i,j,k)\in \bm{y}$ and $T_{ij0}=1$ for all latent spans $(i,j)$ (recall that the $0$-th label is $\emptyset$). Define set $\tilde{\mathcal{\bm{T}}}(\bm{y})$ as all compatible trees with $\bm{y}$. PO-TreeCRF maximizes the total likelihood of all compatible trees:
\begin{align*}
&s(\bm{y}) =\log \sum_{T \in \tilde{\mathcal{\bm{T}}}(\bm{y})} \exp (s(T))\\
&\log p(\bm{y})= s(\bm{y}) -\log Z 
\end{align*}
where $\log Z$ is the log-partition function.
The difficulty is how to estimate $s(\bm{y})$ efficiently. \citet{fu2020nested} propose the masked inside algorithm to tackle this, in which they set all incompatible span (overlapped but not nested with any of $\bm{y}$) values to negative infinity before running the inside algorithm. We refer readers to their paper for more details.

\subsection{Lexicalized Parsing}

Figure \ref{fig:demo_sent} shows an example lexicalized constituency tree. We omit all constituent labels for brevity. Each constituent is annotated by a headword.
A non-leaf constituent span consists of two adjacent sub-constituents and copies the headword from one of them. We refer to the copied headword as the inherited head and the other headword as the non-inherited head. We can draw a dependency arc from the inherited head to the non-inherited head.
A dependency tree can be obtained by reading off all headwords recursively, and hence in this view, a lexicalized constituency tree embeds a  dependency tree and a constituency tree. 

The $O(n^4)$ Eisner-Satta algorithm \cite{eisner-satta-1999-efficient} can be used to calculate the partition function or obtain the best parse if we decompose the score of a lexicalized constituency tree into scores of spans and arcs. We refer interested readers to Appendix \ref{app:eisner_satta} for details of the Eisner-Satta algorithm.  

\section{Model}
\paragraph{Notations}
Given a length-$n$ sentence $\bm{x}={x_0, ..., x_{n-1}}$ with (gold) entity set $\bm{y} := \{(i, j, \bm{\Omega}), \dots\}$, where $(i, j)$ is an unlabeled entity span and $\bm{\Omega}$ is the set of entity labels (there could be multiple labels for one entity). We denote $\bm{\tilde{y}}$ as the set of unlabeled entity spans, i.e., $\bm{\tilde{y}} := \{(i, j), \dots\}$. 

\subsection{Two-stage Strategy and Training Loss} 
The first stage always predicts $2n-1$ spans\footnote{A binary (lexicalized) constituency tree consists of exactly $2n-1$ constituents.} and most of them are not entities. Hence naively adopting the two-stage parsing strategy to nested NER suffers from the imbalanced classification problem when predicting labels in the second stage because the $\emptyset$ label would dominate all the entity labels. To bypass this problem, we modify unlabeled constituency trees by assigning 0-1 labels to 
unlabeled constituency trees, where 0 stands for latent spans and 1 stands for entities. It transfers the burden of identifying non-entities to the first stage, in which the binary classification problem is much more balanced and easier to tackle.
The total training loss can be decomposed into:
\[
L = L_{\text{tree}} + L_{\text{label}} + L_{\text{reg}}
\]
where $L_{\text{tree}}$ is a 0-1 labeled constituency tree loss, $L_{\text{label}}$ is a head-aware labeling loss and $L_{\text{reg}}$ is a regularization loss based on the KL divergence.

\subsection{Stage I: Structure Module} \label{sec:scoring}
\paragraph{Encoding and scoring}
We feed the sentence into the BERT encoder  \cite{devlin-etal-2019-bert}, apply scalar mixing \cite{peters-etal-2018-deep} to the last four layers of BERT, and apply mean-pooling to all subword embeddings to obtain word-level contextual embedding.  We concatenate static word embedding, e.g., GloVe \cite{pennington-etal-2014-glove}, to the contextual embedding to obtain the word representation $a={a_0, .., a_{n-1}}$. Then we feed $a$ into 
a three-layer bidirectional LSTM \cite{hochreiter1997long} network (BiLSTM):
\begin{align}
  & \dots, (\overrightarrow{b_i}, \overleftarrow{b_i}), \dots = \text{BiLSTM}([\dots, a_i, \dots]) \nonumber 
\end{align}
Next, we use deep biaffine scoring functions \cite{DBLP:conf/iclr/DozatM17} to calculate span scores $s^c \in \mathbb{R}^{n \times n \times 2}$ and arc scores $s^d \in \mathbb{R}^{n \times n}$: 
\begin{align*}
   & e^{c,in/out}_i = \text{MLP}^{c,in/out}([\overrightarrow{b_i}; \overleftarrow{b_{i+1}}]) \\
   & e^{d,in/out}_i = \text{MLP}^{d,in/out}([\overrightarrow{b_i}; \overleftarrow{b_{i}}])  \\
   & s^c_{i j}= \text{PN}([e^{c,in}_i;\textbf{1}]^T W^c [e^{c,out}_j;\textbf{1}])  \\
   & s^d_{i j} = \text{PN}([e^{d,in}_i;\textbf{1}]^T W^d [e^{d,out}_j;\textbf{1}]) , 
  \end{align*}
  where MLPs are multi-layer perceptrons that project embeddings into $k$-dimensional spaces;  $W^c \in \mathbb{R}^{(k+1) \times 2 \times (k+1)},W^d \in \mathbb{R}^{(k+1) \times (k+1)}$ are trainable parameters; $\text{PN}$ is Potential Normalization, which normalizes scores to follow unit Gaussian distributions and has been shown beneficial \cite{fu2020nested}.

\paragraph{Scores of trees} A 0-1 labeled lexicalized constituency tree $l$ embeds an unlabeled dependency tree $d$ and a 0-1 labeled constituency tree $c$. The label set is $\{0, 1\}$, where 0 denotes latent spans and 1 denotes entity spans.
  We use a binary rank-3 tensor $C \in R^{n\times n\times 2}$ to represent $c$, where $C_{i j k} = 1$ if and only if there is a span from $x_i$ to $x_j$ with label $k$ in $c$; and a binary matrix $D \in \mathbb{R}^{n \times n}$ to represent $d$, where $D_{i j} = 1$ if and only if there is an arc from $x_i$ to $x_j$ in $d$.  We define the score of $l$ as :
 \begin{align*}
 s(l) &= s(c) + s(d)  \\
  &= \sum_{i j k} C_{i j k} s^c_{i j k} + \sum_{i j} D_{i j} s^d_{i j}
 \end{align*}


\paragraph{Structural tree loss} 
We marginalize all latent spans and arcs out to define the loss:
\begin{align*}
s(\bm{\tilde{y}}) &= \log \sum_{\tilde{T} \in \tilde{\mathcal{T}}} \exp (s(\tilde{T})) \\ 
L_{\text{tree}} &= \log Z - s(\bm{\tilde{y}})
\end{align*}
where $\tilde{\mathcal{T}}$ is the set of all compatible lexicalized trees whose constituents contain $\bm{\tilde{y}}$; $\log Z$ is the log-partition function that can be estimated by the Eisner-Satta algorithm. For each compatible tree $\tilde{T} \in \tilde{\mathcal{T}}$, the 0-1 labels are assigned in accordance with the entity spans in $\bm{\tilde{y}}$. We use a masked version of the Eisner-Satta algorithm (Appendix \ref{app:eisner_satta}) to estimate $s(\bm{\tilde{y}})$.

\paragraph{Regularization loss} \label{sec:soft_constraint}
As previously discussed, entity heads govern only one entity in most cases. But imposing a hard constraint is sub-optimal because there are also cases violating this constraint. Hence we want to encourage the model to satisfy this constraint in a soft manner. Inspired by posterior regularization \cite{ganchev_posterior_2010,zhang-etal-2017-prior}, we build a constrained TreeCRF and minimize the KL divergence between constrained and original unconstrained TreeCRFs. The first problem is how to construct the constrained TreeCRF. We propose to ``hack'' the forward pass (i.e., inside) of the Eisner-Satta algorithm to achieve this: we decrease the arc scores by a constant value (we typically set to 0.4) whenever the parent has already governed an entity during computing the inside values, so it discourages a head having several children and thus governing several spans. We refer readers to Appendix \ref{app:eisner_satta} for more details. The second problem is how to optimize the KL divergence efficiently for exponential numbers of trees. We adopt the specific semiring designed to calculate KL divergences between structured log-linear models \cite{li-eisner-2009-first} from the Torch-Struct library \cite{rush-2020-torch} \footnote{\url{https://github.com/harvardnlp/pytorch-struct/blob/master/torch_struct/semirings/semirings.py}}. The calculation of KL divergence is fully differentiable and thus is amenable to gradient-based optimization methods. It has the same time complexity as the forward pass of the Eisner-Satta algorithm.  We denote the value of KL divergence as $L_{\text{reg}}$.

\subsection{Stage II: Labeling Module}
\label{sec:typing}



To incorporate entity head information when labeling entity spans, we score the assignment of label $l \in \mathcal{L}$ to a span $(i,j)$ with head $x_k$ as follows:
\begin{align}
    & e_i^{l,in/out} = \text{MLP}^{l,in/out}([\overrightarrow{b_i}; \overleftarrow{b_{i+1}}])\nonumber \\
    & e_i^{l,head} = \text{MLP}^{l,head}([\overrightarrow{b_i}; \overleftarrow{b_{i}}])\nonumber \\
    & s^{label}_{i j k l} = \text{TriAff}(e_i^{l,in}, e_j^{l,out}, e_k^{l,head})\nonumber,
\end{align}
where \text{Triaff} is the triaffine scoring function \cite{zhang-etal-2020-efficient}; $\mathcal{L}$ is the set of all labels. We reuse the encoder (BiLSTM) from Stage I.

Nested named entities could have multiple labels. For instance, 7\% entity spans in the NNE dataset \cite{ringland-etal-2019-nne} have multiple labels. We
use a multilabel loss introduced by  \citet{sujianlin_loss}. For each $(i, j, \bm{\Omega}) \in \bm{y}$, consider a potential head $x_k$ with $i \le k \le j$, we define the loss as:
\begin{align*} 
    & l(i, j, k, \bm{\Omega}) = \begin{aligned}[t]
    &\log(1+\sum_{l \in \mathcal{L}/\bm{\Omega}}\exp(s^{label}_{i j k l})) \\
    &+ \log (1+\sum_{l \in \bm{\Omega}}\exp(-s^{label}_{i j k l}))
    \end{aligned}
\end{align*}

Since the gold entity heads are not given, we define the \textbf{head-aware labeling loss} based on expectation over the headword for each entity span:
\[
    L_{\text{label}} = \sum_{(i, j, \bm{\Omega}) \in \bm{y}} \sum_{i \le k \le j} \alpha_{i j k}  l(i,j,k, \bm{\Omega})
\]
where $\alpha_{i j k}$ is the marginal likelihood of $x_k$ being the headword of span $(i, j)$ under the TreeCRF, which satisfies $\sum_{i \le k \le j} \alpha_{i j k} = 1$ and can be estimated efficiently via the backward pass (i.e., backpropagation  \cite{eisner-2016-inside}) of the Eisner-Satta algorithm.

\section{Experiment}

\subsection{Setup}

We conduct experiments on four datasets: ACE2004 \cite{doddington-etal-2004-automatic}, ACE2005 \cite{walker_christopher_ace_2006}, GENIA \cite{kim_genia_2003} and NNE \cite{ringland-etal-2019-nne}. For ACE2004, ACE2005 and GENIA, we use the same data splitting and preprocessing as in \citet{shibuya-hovy-2020-nested}\footnote{\url{https://github.com/yahshibu/nested-ner-tacl2020-transformers}}. For NNE, we use the official preprocessing script\footnote{\url{https://github.com/nickyringland/nested_named_entities/tree/master/ACL2019\%20Paper}} to split train/dev/test sets. We refer readers to  Appendeix \ref{app:impl_detail} for implementation details and to Appendix \ref{app:stat}  for data statistics of each dataset. We report span-level labeled precision (P), labeled recall (R) and labeled F1 scores (F1). We select models according to the performance on development sets. All results are averaged over three runs with different random seeds.

\subsection{Main Result}

We show the comparison of various methods on ACE2004, ACE2005 and GENIA in Table \ref{tab:main}. We note that there is an inconsistency in the data prepossessing. For instance, the data statistics shown in Table 1 of \cite{shibuya-hovy-2020-nested} and Table 5 of \cite{shen-etal-2021-locate} do not match. More seriously, we find \citet{shen-etal-2021-locate,tan2021sequencetoset} use context sentences, which plays a crucial role in their performance improvement but is not standard practice in other work. In addition, they report the best result instead of the mean result. Hence we rerun the open-sourced codes of \citet{shen-etal-2021-locate, tan2021sequencetoset} using our preprocessed data and no context sentences and we report their mean results over three different runs. 
We also rerun the code of PO-TreeCRF for a fair comparison. 

We can see that our method outperforms PO-TreeCRF, our main baseline, by 0.30/2.42/0.64 F1 scores on the three datasets, respectively. Our method has 87.90 and 86.91 F1 scores on ACE2004 and ACE2005, achieving the state-of-the-art performances. On GENIA, our method achieves competitive performance. 

We also evaluate our method on the NNE dataset, whereby there are many multilabeled entities. Table \ref{tab:main_nne} shows the result: our method outperforms Pyramid by 0.27 F1 score.


\begin{table*}[tb!]
\centering
\begin{tabular}{l|ccc|ccc|ccc}
\toprule
\multirow{2}{*}{\textbf{Model}} & \multicolumn{3}{c|}{\textbf{ACE2004}} & \multicolumn{3}{c|}{\textbf{ACE2005}} & \multicolumn{3}{c}{\textbf{GENIA}} \\
                                 & P & R & F1 & P & R & F1 & P & R & F1 \\\hline\multicolumn{10}{l}{\textbf{Comparable}} \\
SH & - & - & -  & 83.30  & 84.69  & 83.99   & 77.46 & 76.65 & 77.05   \\
Pyramid-Basic   &  86.08 & 86.48  &  86.28  &  83.95 & 85.39  & 84.66   &  78.45 & 78.94  & 79.19   \\
W(max)  & 86.27  & 85.09 &  85.68  & 85.28 & 84.15  &  84.71  & 79.20  & 78.16  &  78.67  \\
PO-TreeCRFs$^\dagger$  & 87.62 & 87.57 & 87.60 & 83.34  & 85.67  & 84.49& 79.10  & 76.53  & 77.80   \\
Seq2set$^\dagger$     & 87.05  & 86.26  &  86.65  & 83.92  & 84.75 & 84.33  & 78.33  & 76.66  & 77.48  \\
Locate\&Label$^\dagger$ & 87.27 & 86.61 & 86.94 & 86.02 & 85.62 & 85.82 & 76.80 & 79.02 & 77.89 \\
BARTNER & 87.27 & 86.41 & 86.84 & 83.16 & 86.38 & 84.74 & 78.57 & 79.3 & 78.93 \\
Ours  &  87.39 & 88.40  & 87.90  & 85.97  & 87.87 & 86.91 & 78.39 & 78.50 & 78.44 \\\hline
\multicolumn{10}{l}{\textbf{For reference}} \\
SH\hfill[F] & - & - & -  & 83.83  & 84.87  & 84.34 & 77.81 & 76.94 & 77.36 \\
Pyramid-Full \hfill[A]  &  87.71 & 87.78  &  87.74  &  85.30 & 87.40 & 86.34   &  - & -  & -  \\
PO-TreeCRFs\hfill[D]          &  86.7 & 86.5  &  86.6  & 84.5  & 86.4  & 85.4 & 78.2  & 78.2 & 78.2   \\
Seq2set\hfill[C,P,D]  &  88.46 & 86.10  &  87.26  &  87.48 & 86.63 & 87.05   &  82.31 & 78.66  & 80.44  \\ 
Locate\&Label\hfill[C,P,D]  &  87.44 & 87.38  &  87.41  & 86.09 & 87.27 & 86.67   & 80.19 & 80.89  & 80.54  \\ 
\bottomrule
\end{tabular}
\caption{Results on ACE2004, ACE2005 and GENIA. SH: \citet{shibuya-hovy-2020-nested}; Pyramid-Basic/Full: \citet{wang-etal-2020-pyramid}\protect\footnotemark; W(max/logsumexp): \citet{wang-etal-2021-nested}\protect\footnotemark; PO-TreeCRFs: \citet{fu2020nested}; Seq2set: \citet{tan2021sequencetoset}
\label{tab:main}; Locate\&Label: \citet{shen-etal-2021-locate}; BARTNER: \citet{yan-etal-2021-unified-generative}. Labels in square brackets stand for the reasons of the results being incomparable to ours. F: +Flair; A: +ALBERT, C: context sentences, P: POS tags, D: different data preprocessing. $\dagger$ denotes that we rerun their open-sourced codes using our data.
}
\end{table*}
\addtocounter{footnote}{-2}



\begin{table}[tb!]
\centering
\begin{tabular}{l|ccc}
\toprule
\multirow{2}{*}{\textbf{Model}} & \multicolumn{3}{c}{\textbf{NNE}}\\
& P & R & F1 \\\hline
Pyramid-Basic &  93.97 & 94.79  &  94.37  \\
Ours & 94.32  & 94.97  & 94.64   \\
\bottomrule
\end{tabular}
\caption{Results on NNE.}
\label{tab:main_nne}
\end{table}

\section{Analysis}

\subsection{Ablation Studies}\label{sec:ablation_struct}
We conduct a thorough ablation study of our model on the ACE2005 test set.  Table \ref{tab:abl} shows the result.

\paragraph{Structured vs. unstructured} We study the effect of structural training and structured decoding as a whole. ``Unstructured'' is a baseline that adopts the local span classification loss and local greedy decoding. ``1-stage`` is our re-implementation of PO-TreeCRF, which adopts the latent structural constituency tree loss and uses the CYK algorithm for decoding. ``1-stage+\texttt{LEX}'' adopts the latent structural lexicalized constituency tree loss and uses the Eisner-Satta algorithm for decoding. All methods use the same neural encoders. We can see that ``1-stage'' outperforms the unstructured baseline by 0.33 F1 score. Further, ``1-stage+\texttt{LEX}'' outperforms ``1-stage'' by 0.25 F1 score, verifying the effectiveness of using latent lexicalized constituency tree structures.

\paragraph{1-stage vs. 2-stage} On the unstructured model, we adopt a 0-1 local span classification loss in the first stage of the two-stage version, and we observe that the two-stage version performs similarly the one-stage version. On the other hand, we observe improvements on structured methods: ``2-stage'' outperforms ``1-stage'' by 0.23 F1 score and ``2-stage+\texttt{LEX}'' outperforms ``1-stage+\texttt{LEX}'' by 0.18 F1 scores, validating the benefit of adopting the two-stage strategy. Moreover, "2-stage(0/1)+\texttt{LEX}" outperforms "2-stage+\texttt{LEX}" by 0.15 F1 score, suggesting the effectiveness of bypassing the imbalanced classification problem.

\stepcounter{footnote}\footnotetext{They did not report Pyramid-Full with BERT only. However, with BERT+ALBERT, Pyramid-Full only outperforms Pyramid-Basic with a small margin ($<0.1$). }
\stepcounter{footnote}\footnotetext{The \textit{max} and \textit{logsumexp} versions are the best models for BERT only and BERT+Flair respectively.}

\paragraph{Effect of structural training and decoding} We study the importance of structural training and decoding in a decoupled way here. ``-parsing'' denotes the case that we use the latent lexicalized constituency tree loss for training, while we do not use the Eisner-Satta algorithm for parsing and instead predict spans locally whenever their label score of 1 is greater than that of 0. 
We can see that it causes a performance drop of 0.49 F1 score, indicating the importance of structural decoding, i.e., parsing. It is also worth noting that ``-parsing'' outperforms the unstructured baseline by 0.42 F1 score, showing the benefit of structural training even without structural decoding.  

\paragraph{Effect of head regularization}  We can see that using the regularization loss brings an improvement of 0.24 F1 score (86.32->86.56). In the case study (Section \ref{sec:case_study}), we observe that some common errors are avoided because of this regularization.

\paragraph{Effect of head-aware labeling loss} We can see that using the head-aware labeling loss brings an improvement of 0.30 F1 score (86.32 -> 86.62). When combined with the head regularization, we achieve further improvements because of more accurate head estimation (Appendix \ref{app:head}). 


\begin{table}[tb!]
\begin{tabular}{l|lll}
\toprule
\multicolumn{1}{c|}{\textbf{Model}} &
  \multicolumn{1}{c}{\textbf{P}} &
  \multicolumn{1}{c}{\textbf{R}} &
  \multicolumn{1}{c}{\textbf{F1}} \\ \hline
Unstructured(1-stage) & 83.76 & 87.17 & 85.43 \\
Unstructured(2-stage) & 84.23 & 86.62 & 85.41 \\
1-stage & 84.08 & 87.52 & 85.76 \\
1-stage + \texttt{LEX}    & 84.26 & 87.83 & 86.01 \\
2-stage & 84.68 & 87.33 & 85.99 \\
2-stage + \texttt{LEX}  & 84.60 & 87.80 & 86.17 \\
2-stage (0-1) + \texttt{LEX}  & 84.83 & 87.87 & 86.32 \\
\small{\quad - parsing} & 84.26 & 87.40 & 85.83 \\
\small{\quad +  head regularization} & 85.84 & 87.30 & 86.56 \\
\small{\quad + head-aware labeling}       & 85.50 & 87.77 & 86.62 \\
\small{\quad + both} (our final model)                          & \textbf{85.97} & \textbf{87.87} & \textbf{86.91} \\ \bottomrule 
\end{tabular}
\caption{Ablation studies on the ACE2005 test set. \texttt{LEX} represents lexicalized structures.}
\label{tab:abl}
\end{table}

\newcommand{\cc}[2]{\textcolor{blue}{[}#1\textcolor{blue}{]\textsuperscript{#2}}}
\newcommand{\ww}[2]{\textcolor{red}{[}#1\textcolor{red}{]\textsuperscript{#2}}}

\begin{table*}[tb!]
    \centering
    \begin{tabular}{p{0.14\linewidth} | p{0.82\linewidth}}
    \toprule
    \textbf{Model} & \textbf{Prediction}\\\hline
     2-stage & \cc{\underline{I}}{PER} have never heard of \ww{a pig like \ww{\underline{this}}{WEA}}{WEA} before !\\
  2-stage (0-1)$^\ddagger$ & \cc{\underline{I}}{PER} have never heard of a pig like this before !
 \\ \midrule
    2-stage (0-1) & 
    \textcolor{blue}{[}\underline{Police}\textcolor{blue}{]}\textsuperscript{\textcolor{blue}{PER}} surrounded \textcolor{red}{[}\underline{this} bus near \textcolor{blue}{[}\underline{the} airport\textcolor{blue}{]}\textsuperscript{\textcolor{blue}{FAC}}\textcolor{red}{]}\textsuperscript{\textcolor{red}{VEH,FAC}} with \textcolor{blue}{[}\underline{guns}\textcolor{blue}{]}\textsuperscript{\textcolor{blue}{WEA}} drawn . \\
     + both$^\ddagger$ & \textcolor{blue}{[}\underline{Police}\textcolor{blue}{]}\textsuperscript{\textcolor{blue}{PER}} surrounded \textcolor{blue}{[}this \underline{bus}\textcolor{blue}{]}\textsuperscript{\textcolor{blue}{VEH}} near \textcolor{blue}{[}the \underline{airport}\textcolor{blue}{]}\textsuperscript{\textcolor{blue}{FAC}} with \textcolor{blue}{[}\underline{guns}\textcolor{blue}{]}\textsuperscript{\textcolor{blue}{WEA}} drawn . \\[0.25ex]\cdashline{1-2}\noalign{\vskip 0.25ex}
     + both$^\ddagger$ & \cc{\underline{Blix}}{PER} stressed that \cc{\underline{council}}{ORG} resolutions call for \cc{\cc{\underline{U.N.}}{ORG} \underline{inspectors}}{PER} to have access to \cc{all sites and people \underline{in} \cc{\underline{Ira\smash{q}}}{GPE}}{FAC,PER} .
 \\ 
    \bottomrule
    \end{tabular}
    \caption{Two sentences with predicted entity decorated. \textcolor{blue}{Blue} entities are correct and \textcolor{red}{red} entities are wrong. The underlined \underline{words} are the entity heads. Models annotated with $^\ddagger$ predict all entities correctly.}
    \label{tab:case_study}
\end{table*}

\begin{table*}[tb]
\centering
\begin{tabular}{c|c}
\toprule
\textbf{Type} & \textbf{Most Frequent Headwords}                                         \\ \hline
PER           & you, I, he, they, i, his, of, their, we, who                              \\
LOC           & world, \textcolor{red}{of}, area, there, coast, \textcolor{red}{where}, \textcolor{red}{beach}, desert, \textcolor{red}{Southeast}, that      \\
ORG           & we, they, Starbucks, its, court, company, military, of, their, companies  \\
GPE           & U.S., Indonesia, Baghdad, city, state, Russian, we, country, Iraqi, where \\
FAC           & airport, house, jail, \textcolor{red}{in}, prison, street, \textcolor{red}{of}, it, hospital, home          \\
VEH           & \textcolor{red}{of}, car, \textcolor{red}{in}, aircraft, that, bus, plane, lincoln, deck, its               \\
WEA & gun, weapons, arms, guns, firearms, missile, bullet, knife, rifles, Kalashnikov \\ \bottomrule
\end{tabular}
\caption{The most common (top 10) headwords of each entity type predicted by our method  on the ACE2005 test set. \textcolor{red}{Red} words are not headwords in the gold annotation. }
\label{tab:show_head}
\end{table*}



\begin{table}[tb!]
\centering
\begin{tabular}{l|cc}
\toprule
 Model  & Train    &  Sents/sec     \\ \hline
PO-TreeCRF & 2m1s & 205 \\
2-stage & 2m15s & 184 \\
2-stage + \texttt{LEX}  & 2m23s & 173 \\ \hline
Seq2set & 3m24s & 122\\
Locate\&Label & 4m23s & 94
\\ \bottomrule
\end{tabular}
\caption{Speed comparison for training one epoch on ACE2005. }
\label{tab:speed}
\end{table}

\subsection{Case Study}
\label{sec:case_study}


Table \ref{tab:case_study} shows example predictions of our models. In the first pair, ``2-stage'' predict reasonable structures (visualized in \ref{app:parse_result}), but fail to label entities, whereas ``2-stage (0-1)'' predicts further correct labels. The second pair shows that, by constraining head sharing and head-aware entity labeling, ``+both'' successfully detect \text{bus} as a headword, then produce correct entity boundaries and labels. Besides, ``+both'' can be seen to handle both fine-grained and coarse-grained entities in the last two predictions: \textit{this bus near the airport} is predicted into two entities but \textit{all sites and people in Iraq} remains one multilabeled entity.

Table \ref{tab:show_head} gives the most common headwords of each type predicted by our model on ACE2005. We find that the most frequently predicted headwords are gold headwords\footnote{ACE2005 is additionally annotated with headwords. We only use them for evaluation.}, except for some common function words, e.g., \textit{in} and \textit{of}. It proves the ability of our model in recognizing headwords.


\interfootnotelinepenalty=10000
\subsection{Speed Comparison}
One concern regarding our method is that since the Eisner-Satta algorithm has a $O(n^4)$ time complexity, it would be too slow to use for NER practitioners. Fortunately, the Eisner-Satta algorithm is amenable to highly-parallelized implementation so that $O(n^3)$ out of $O(n^4)$ can be computed in parallel \cite{zhang-etal-2020-efficient,rush-2020-torch}, which greatly accelerates parsing. We adapt the fast implementation of \citet{Yang2022Combining} \footnote{\url{https://github.com/sustcsonglin/span-based-dependency-parsing/blob/main/src/inside/eisner_satta.py}}. Empirically, we  observe linear running time on GPUs in most cases. We show the comparison of (both training and decoding) running time in Table \ref{tab:speed}. We measure the time on a machine with Intel Xeon Gold 6278C CPU and NVIDIA V100 GPU.

We can see that compared with PO-TreeCRF, which also uses a highly-parallelized implementation of the $O(n^3)$ CYK algorithm, our method is around 20\% slower in training and decoding, which is acceptable. Notably, both PO-TreeCRF and our method are much faster than Seq2Set \cite{tan2021sequencetoset} and Locate\&Label\cite{shen-etal-2021-locate}.


\section{Related Work}

\paragraph{Nested NER}
Nested NER has been receiving increasing attentions and there are many methods proposed to tackle it. We roughly categorize the methods into the following groups: (1) Span-based methods: \citet{luan-etal-2019-general,yu-etal-2020-named, li-etal-2021-span} directly assign scores to each potential entity span. (2) Layered methods: \citet{ju-etal-2018-neural,fisher-vlachos-2019-merge} dynamically merge subspans to larger spans and
\citet{shibuya-hovy-2020-nested, wang-etal-2021-nested} use linear-chain CRFs and recursively find second-best paths for predicting nested entities.  (3) Hypergraph-based methods: \citet{lu-roth-2015-joint, katiyar-cardie-2018-nested} propose different hypergraph structures to model nested entities but suffer from the spurious structure issue, and \citet{wang-lu-2018-neural} solve this issue later.  (4) Object-detection-based methods: \citet{shen-etal-2021-locate} adapt classical two-stage object detectors to tackle nested NER and \citet{tan2021sequencetoset} borrow the idea from DETR \cite{DBLP:conf/eccv/CarionMSUKZ20}. (5) Parsing-based methods \cite{finkel-manning-2009-nested,wang-etal-2018-neural-transition, fu2020nested, Yang2022Pointer}. (6) Sequence-to-sequence methods \cite{yan-etal-2021-unified-generative}. 

Our method belongs to parsing-based methods. \citet{finkel-manning-2009-nested} use a non-neural TreeCRF parser. \citet{wang-etal-2018-neural-transition} adapt a shift-reduce transition-based  parser. \citet{fu2020nested} use a span-based neural TreeCRF parser. Recently, \citet{Yang2022Pointer} propose a bottom-up constituency parser with pointer networks to tackle nested NER as well. All of them cast nested NER to constituency parsing, while we cast nested NER to lexicalized constituency parsing and our method is thus able to model entity heads.




\paragraph{Structured models using partial trees}
Full gold parse trees are expensive to obtain, so there are many methods proposed to marginalize over latent parts of partial trees, performing either approximate marginalization via loopy belief propagation or other approximate algorithms \cite{naradowsky2012improving, durrett2014joint} or exact marginalization via dynamic programming algorithms \cite{li-etal-2016-active,zhang-etal-2020-efficient,fu2020nested,Zhang2021SemanticRL}.
\citet{naradowsky2012improving, durrett2014joint} construct factor graph representations of syntactically-coupled NLP tasks whose structures can be viewed as latent dependency or constituency trees, such as NER, semantic role labeling (SRL), and relation extraction. \citet{li-etal-2016-active, zhang-etal-2020-efficient} perform partial marginalization to train (second-order) TreeCRF parsers for partially-annotated dependency parsing. \citet{Zhang2021SemanticRL} view arcs in SRL as partially-observed dependency trees; \citet{fu2020nested} view entities in nested NER as partially-observed constituency trees; and we view entities in nested NER as partially-observed lexicalized constituency trees in this work. 


\paragraph{Lexicalized parsing}
Probabilistic context-free grammars (PCFGs) have been widely used in syntactic parsing. Lexicalized PCFGs (L-PCFGs) leverage headword information to disambiguate parsing and are thus more expressive. \citet{eisner-satta-1999-efficient} propose an efficient $O(n^4)$ algorithm for lexicalized parsing. \citet{collins-2003-head} conduct a thorough study of lexicalized parsing. Recently, neurally parameterized L-PCFGs have been used in unsupervised joint dependency and constituency parsing \cite{zhu-etal-2020-return, yang-etal-2021-neural}. 
Our work removes the grammar components and adapts the dynamic programming algorithm of lexicalized parsing \cite{eisner-satta-1999-efficient} in the spirit of span-based constituency parsing \cite{stern-etal-2017-minimal}.



\section{Conclusion}
We have presented a parsing-based method for nested NER, viewing entities as partially-observed lexicalized constituency trees, motivated by the close relationship between entity heads and entity recognition. Benefiting from structural modeling, our model does not suffer from error propagation and heuristic head choosing and is easy for regularizing predictions. Furthermore, our highly-parallelized implementation enables fast training and inference on GPUs. Experiments on four benchmark datasets validate the effectiveness and efficiency of our proposed method.

\section*{Acknowledgments}
We thank the anonymous reviewers for their constructive comments. This work was supported by the National Natural Science Foundation of China (61976139).

\bibliography{anthology,custom}
\bibliographystyle{acl_natbib}

\appendix

\section{Details of the Eisner-Satta algorithm}
\label{app:eisner_satta}

Table \ref{tab:eisner-satta} describes the Eisner-Satta algorithm in the parsing-as-deduction framework. Each deductive rule of the Eisner-Satta algorithm has only one word participating in the computation in addition, e.g., $p$ and $h$, resulting in one-order higher than the CYK algorithm.

The masked version of the Eisner-Satta algorithm masks scores similar to PO-TreeCRF except for different label sets in our model ``2-stage'' and ``2-stage (0/1)''. For the construction of constrained trees, we introduce a minor penalty (0.4 in our paper) on type I items' scores if the item represents a gold entity. We show the pseudocode of the standard Eisner-Satta algorithm, the masked version of the Esiner-Satta algorithm and the construction of constrained trees all in Algorithm \ref{alg:eisner_satta}.
\begin{table}[tb]
	\centering
	{\setlength{\tabcolsep}{.0em}
		\begin{tabular}{r}
			\toprule 
			\begin{minipage}{\linewidth}
    \textit{Items}: 
    \renewcommand{\labelenumi}{\Roman{enumi}}
    \begin{enumerate}
      \item  $[i, j, h, -]$: span $[i, j]$ is headed by word $w_h$ and its parent is not determined. $i \le h \le j$.
      \item $[i, j, -, p]$: span $[i, j]$ is headed by arbitrary word $w_h$. The common parent is $w_p$. $p<i$ or $k<p$.
    \end{enumerate}
    \textit{Axiom items}: $[i, i, i, -]$, $1 \le i \le n$ \\ 
    \textit{Goal items}: $[1, n, r, -]$, $1 \le r \le n$ \\ 
    \textit{Deductive rules:} 
    \begin{enumerate}
        \item \phantom{} \infer[\text{attach left/right}]{[i, k, -, p]}{[i, k, h, -]} 
        \item \phantom{} \infer[\text{complete left}]{[i, k, p, -]}{[i, j, -, p] & [j+1, k, p, -]} 
        \item \phantom{} \infer[\text{complete right}]{[i, k, p, -]}{[i, j, p, -]  & [j+1, k, -, p]}
    \end{enumerate}
			\end{minipage}\\
			\bottomrule
	\end{tabular}}
	\caption{The Eisner-Satta algorithm described in the parsing-as-deduction framework. }
	\label{tab:eisner-satta}
\end{table}

\section{Experiments}
\label{sec:appendix}

\subsection{Implementation Details}
\label{app:impl_detail}
We use BERT (bert-large-cased) and GloVe (6B-100d) to obtain word representations for ACE2004, ACE2005, and NNE. For GENIA, we use BioBERT (biobert-large-cased-v1.1) \cite{10.1093/bioinformatics/btz682} and BioWordvec \cite{chiu-etal-2016-train} instead to match its domain. 
The hidden size of BiLSTM is set to 400.
We use an Adam optimizer \cite{DBLP:journals/corr/KingmaB14} and a linear learning rate scheduler. We warm up training for 2 epochs and decay learning rates to 0 linearly for the rest of the epochs. The peak learning rates are $5e-5$ for BERT/BioBERT and $5e-3$ for the other parts of the neural networks. 

\begin{table}[tb!]
    \centering
\resizebox{\linewidth}{!}{
\begin{tabular}{l|ccccccc}\toprule
 &
  \multicolumn{1}{l}{PER} &
  \multicolumn{1}{l}{LOC} &
  \multicolumn{1}{l}{ORG} &
  \multicolumn{1}{l}{GPE} &
  \multicolumn{1}{l}{FAC} &
  \multicolumn{1}{l}{VEH} &
  \multicolumn{1}{l}{WEA} \\\hline
$\rho$ & 0.57 & 0.02 & 0.18 & 0.14 & 0.05 & 0.03 & 0.02 \\\hline
PER  & 0.92 & 0.00 & 0.06 & 0.03 & 0.01 & 0.03 & 0.00 \\
LOC  & 0.00 & 0.74 & 0.00 & 0.02 & 0.01 & 0.01 & 0.00 \\
ORG  & 0.02 & 0.00 & 0.83 & 0.02 & 0.03 & 0.02 & 0.00 \\
GPE  & 0.00 & 0.07 & 0.03 & 0.87 & 0.04 & 0.00 & 0.00 \\
FAC  & 0.00 & 0.06 & 0.01 & 0.00 & 0.77 & 0.04 & 0.00 \\
VEH  & 0.00 & 0.00 & 0.00 & 0.00 & 0.01 & 0.73 & 0.00 \\
WEA  & 0.00 & 0.00 & 0.00 & 0.00 & 0.01 & 0.00 & 0.90 \\
$\emptyset$ & 0.06 & 0.13 & 0.08 & 0.06 & 0.12 & 0.18 & 0.10 \\
\bottomrule
\end{tabular}
}
\caption{Error distribution on the ACE2005 test set normalized along with columns. $\rho$ is the gold label distribution. Each row is a gold label and each column is a predicted label. $\emptyset$ denotes entities not recognized by our model.}
\label{tab:by_cat}
\end{table}

\subsection{Data statistics}
\label{app:stat}
Table \ref{tab:stat} shows the statistics of ACE2004, ACE2005, GENIA and NNE. We report the number of multilabeled entities and single-word entities in addition. 

\begin{table*}[tb!]
\centering
\small
\begin{tabular}{l|ccc|ccc|ccc|ccc}
\toprule
 & \multicolumn{3}{c|}{\textbf{ACE2004}} & \multicolumn{3}{c|}{\textbf{ACE2005}} & \multicolumn{3}{c|}{\textbf{GENIA}} & \multicolumn{3}{c}{\textbf{NNE}} \\
              & train & dev  & test & train & dev  & test & train & dev  & test & train  & dev   & test  \\ \hline
\# sentences  & 6198  & 742  & 809  & 7285  & 968  & 1058 & 15022 & 1669 & 1855 & 43457  & 1989  & 3762  \\
- nested      & 2718  & 294  & 388  & 2797  & 352  & 339  & 3222  & 328  & 448  & 28606  & 1292  & 2489  \\ \hline
\# entities   & 22195 & 2514 & 3034 & 24827 & 3234 & 3041 & 47006 & 4461 & 5596 & 248136 & 10463 & 21196 \\
- nested      & 10157 & 1092 & 1417 & 9946  & 1191 & 1179 & 8382  & 818  & 1212 & 206618 & 8487  & 17670 \\
- single-word &   11527    &  1363    & 1553     &  13988     & 1852     & 1706     &  12933     & 1009     &  1392    &   166183     &  7291     & 14397      \\
- multi-type  &   3    &   1   &  1    &   9    &  3    &    2  &   21    &  5    &  5    &  16769      & 792      &  1583     \\ \bottomrule
\end{tabular}
\caption{Statistics of ACE2004, ACE2005, GENIA and NNE. An entity is considered nested if contains any entity or is contained by any entity. A sentence is considered nested if contains any nested entity.}
\label{tab:stat}
\end{table*}

\subsection{Studies on Headwords}
\label{app:head}

We conduct more experiments to analyze the behavior of head regularization. Table \ref{tab:diff_c} shows the results of models trained with different penalty constants of the head regularization. $c=0$ means no constraint applied, and larger $c$ means harder constraint. We observe that too hard constraints (e.g., $c=1$) are less effective than proper constraints (e.g., $c=0.4$). We choose $c=0.4$ as the penalty constant for experiments in the main body. Table \ref{tab:bias_decoding} shows the results if we apply head regularization only when decoding. We observe that the overall performance changes marginally, although the number of shared heads is significantly reduced, possibly because the head accuracy is still low and the labeling module is trained to pay less attention to the headwords as they are noisy. Finally, we analyze the number of shared heads and the head accuracy for models trained with head regularization and head-aware entity labeling. Table \ref{tab:bias_training} shows few shared heads and high head accuracy, consistent with the high overall performance. Besides, we observe that adding the head-aware entity labeling does not reduce the shared headwords much, showing the limitation of models to learn such prior knowledge. 
 
\begin{table*}[tb!]
    \centering 
    \begin{tabular}{c|cccccccc} \toprule
         $c$  & 0 & 0.1 & 0.2 & 0.3 & 0.4 & 0.5 & 0.6\\\hline
        F1 & 86.32 & 86.45 & 86.54 & 86.53 & 86.56 & 86.49 & 86.41\\
        \bottomrule
    \end{tabular}
    \caption{The impact of different constants used to construct constrained trees for training on ACE2005. A higher value means harder constraints.}
    \label{tab:diff_c}
\end{table*}

\begin{table*}[tb!]
\centering
\begin{tabular}{c|cccccc}
\toprule
$c$            & -2    & 0     & 0.2   & 0.4   & 0.6   & 1     \\ \hline
F1          & 86.38 & 86.44 & 86.46 & 86.46 & 86.43 & 86.41 \\
\#shared & 347   & 234   & 30    & 10    & 7     & 6     \\
Head acc.   & 43.19 & 48.45 & 57.27 & 57.94 & 58.33 & 58.08 \\ \bottomrule
\end{tabular}
\caption{Results of different constants when decoding. \#shared denotes the number of entities having shared headwords. Models are trained without the head regularization. Head accuracy do not count single word spans. Results are of one run. }
\label{tab:bias_decoding}
\end{table*}

\begin{table*}[tb!]
\centering
\begin{tabular}{c|cccccc}
\toprule
               & 0      & 0.4  & 0 + HA  & 0.4 + HA      \\ \hline
\#shared  & 234      & 73    & 216  & 10      \\
Head acc. & 48.45  & 59.42 & 73.58 & 81.00 \\ \bottomrule 
\end{tabular}
\caption{Number of shared heads and head accuracy on the ACE2005 test set. HA means head-aware entity labeling. The head accuracy do not count single word spans. Results are of one run. }
\label{tab:bias_training}
\end{table*}

\subsection{Error Distribution}
We report the error distribution in Table \ref{tab:by_cat}. Compared with PO-TreeCRF, we reduce the error rates off all extremely imbalanced classes (VEH, FAC, LOC and WEA).

\subsection{Predicted Parse Tree}
\label{app:parse_result}

\begin{figure*}[tb!]
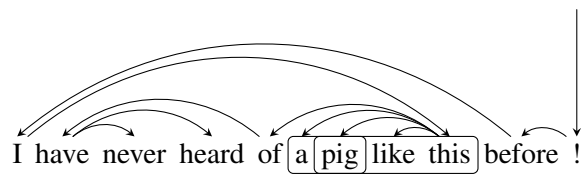
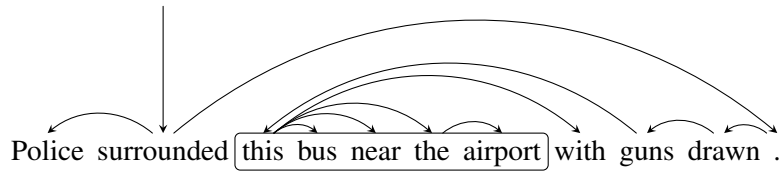
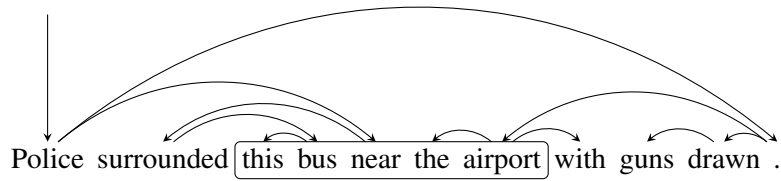

  \centering
  \begin{subfigure}{\linewidth}
  \centering
  \begin{dependency}[hide label,arc edge,arc angle = 40]
    \begin{deptext}
      I \& have \& never \& heard \& of \& a \& pig \& like \& this \& before \& ! \\
    \end{deptext}
    \depedge{10}{1}{}
    \depedge{5}{2}{}
    \depedge{2}{3}{}
    \depedge{2}{4}{}
    \depedge{9}{5}{}
    \depedge{9}{6}{}
    \depedge{9}{7}{}
    \depedge{9}{8}{}
    \depedge{1}{9}{}
    \depedge{11}{10}{}
    \deproot    {11}{}
    \wordgroup{1}{6}{9}{a0}
    \wordgroup{1}{7}{7}{a1}
  \end{dependency}
  \caption{A tree predicted by ``2-stage''. It produce reasonable structures, but the labeling module can not label them well.}
  \label{fig:case_correct_struct}
  \end{subfigure}
  \begin{subfigure}{\linewidth}
    \centering
    \begin{dependency}[hide label,arc edge,arc angle = 40]
      \begin{deptext}
        Police \& surrounded \& this \& bus \& near \& the  \& airport \& with \& guns \& drawn \& . \\
      \end{deptext}
      \depedge{2}{1}{}
      \deproot   {2}{}
      \depedge{9}{3}{}
      \depedge{3}{4}{}
      \depedge{3}{5}{}
      \depedge{3}{6}{}
      \depedge{6}{7}{}
      \depedge{3}{8}{}
      \depedge{10}{9}{}
      \depedge{11}{10}{}
      \depedge{2}{11}{}
      \wordgroup{1}{3}{7}{a0}
    \end{dependency}
    \caption{A tree predicted by ``2stage (0-1)''. It fails to detect ``bus'' and ``airport'' as headwords. }
    \label{fig:case_wrong_struct}
    \end{subfigure}
  \begin{subfigure}{\linewidth}
    \centering
    \begin{dependency}[hide label,arc edge,arc angle = 40]
      \begin{deptext}
        Police \& surrounded \& this \& bus \& near \& the  \& airport \& with \& guns \& drawn \& . \\
      \end{deptext}
      \deproot   {1}{}
      \depedge{5}{2}{}
      \depedge{4}{3}{}
      \depedge{2}{4}{}
      \depedge{1}{5}{}
      \depedge{7}{6}{}
      \depedge{11}{7}{}
      \depedge{7}{8}{}
      \depedge{10}{9}{}
      \depedge{11}{10}{}
      \depedge{1}{11}{}
      \wordgroup{1}{3}{7}{a0}
    \end{dependency}
    \caption{A tree predicted by ``2-stage (0-1) + both''. It detect ``bus'' and ``airport'' as headwords correctly. The span \textit{this bus near the airport} do not exist on the tree. }
     \label{fig:correct_struct2}
    \end{subfigure}
  \caption{Predicted dependency trees. We highlight interesting spans.}
  \label{fig:trivial_sol}
\end{figure*}

Here we draw the parse trees in \ref{sec:case_study}. Fig. \ref{fig:case_correct_struct} shows a tree produced by ``2-stage'', which is reasonable. But the label module of ``2-stage'' fail to label spans correctly due to the label imbalance problem. ``2-stage (0-1)'' predict the same tree but correct labels. Fig. \ref{fig:case_wrong_struct} shows a tree predicted by ``2-stage (0-1)''. The model fail to detect headwords, e.g., \textit{bus} and \textit{airport}. In contrast, Fig. \ref{fig:correct_struct2} shows a tree predicted by ``2-stage (0-1) + both'', in which shared heads are much fewer and correct headwords are found.

\begin{algorithm*}
\small
\newcommand{\nonl}{\renewcommand{\nl}{\let\nl\oldnl}}
\SetAlgoLined
\DontPrintSemicolon
\SetKwInput{KwDefine}{define}
\SetKwInput{KwInitialize}{initialize}
\SetKwInput{KwInput}{input} 
\KwInput{$s_c \in \mathbb{R}^{n \times n  \times B}$ for span scores, where $B$ is \#sent in a batch}
\KwInput{$s_d \in \mathbb{R}^{n \times n  \times B}$ for arc scores}
\KwInput{$enable\_soft\_constraint$ for whether enable the soft exclusive head constraint}
\KwInput{$mask\in \mathbb{R}^{n \times n}$ for incompatible spans. (optional)}
\KwDefine{$H \in \mathbb{R}^{n \times n \times n \times B}$ for type I span in Table \ref{tab:eisner-satta}}
\KwDefine{$P \in \mathbb{R}^{n \times n \times n \times B}$ for type II span in Table \ref{tab:eisner-satta}}
\KwInitialize{$H_{:,:,:}=-\infty, P_{:,:,:}=-\infty$}
\If{$mask$ is given}{
    for all $i,j$, $s_c[i,j]=-\infty$ if $mask[i,j]$ is true.
}
\For{$i=0$ \KwTo $n-1$}{
 $H[i,i,i] = s_c[i,i]$
 \For{$j=0$ \KwTo $n-1$}{
 $P[i,i,j] = s_d[i,j] + H[i,i,i]$ 
 }
 \If{$enable\_soft\_constraint$}{
     $H[i,i,i] -= c$
    \tcp*{$c$ is a small positive constant (0.4 in our paper).}
    
    \tcp*{Equivalent to minus $c$ for arcs headed by $i$.}
}
}
\For{$w=1$ \KwTo $n-1$}{
\For{$i=0$ \KwTo $n-w-1$}{
    $j=i+w$
    \For{$h=i$ \KwTo $j$ }{
    $H[i,j,h] = s_c[i,j] +
      \log\sum\limits_{r\in[i,j]}[\exp(P[i,r,h] + H[r+1, j,h]) +\exp(H[i,r,h] + P[r+1, j,h])]$
      
            \tcp*{complete left/right}
      } 
  \For{$p=0$ \KwTo $n-1$}{
    $P[i,j,p] = \log\sum\limits_{h\in [i, j]}\exp(H[i,j,h] + s_d[h,p])$
    \tcp*{attach left/right}
}
    \If{$enable\_soft\_constraint$}{
        \For{$h=i$ \KwTo $j$ }{
            $H[i,j,h] -= c$
        }
        }
}
}
\Return{$H[0,n-1,0]\equiv \log Z$}
 \caption{The Eisner-Satta Algorithm}
 \label{alg:eisner_satta}
\end{algorithm*}

\end{document}